\pdfoutput=1
\documentclass[11pt]{article}
\usepackage{coling2020}
\usepackage{times}
\usepackage{url}
\usepackage{latexsym}
\usepackage{subcaption}
\colingfinalcopy 

\usepackage{graphicx}
\usepackage{tabularx}
\usepackage{soul}
\usepackage{color}

\usepackage{epstopdf}

\usepackage{hyperref}
\usepackage{xstring}
\usepackage{xspace}
\newcommand{\eat}[1]{}
\usepackage{times}
\usepackage{latexsym}
\usepackage{multirow}
\usepackage{multicol}
\usepackage{rotating}
\usepackage{xspace}
\usepackage{url}
\usepackage{footnote}
\newcommand{\namep}{{PerSenT}\xspace}

\usepackage{hyperref}

\title{Author's Sentiment Prediction}

\author{
  Mohaddeseh Bastan, Mahnaz Koupaee, Youngseo Son, Richard Sicoli, \\ \textbf{Niranjan Balasubramanian}
 \\
  Stony Brook University, Stony Brook, New York \\
  {\tt \{mbastan, mkoupaee, yson, rpsicoli,  niranjan\}@cs.stonybrook.edu} 
  \\ }

\date{}

\begin{document}
\maketitle
\begin{abstract}
We introduce \namep, a dataset of crowd-sourced annotations of the sentiment expressed by the authors towards the main entities in news articles. The dataset also includes paragraph-level sentiment annotations to provide more fine-grained supervision for the task. Our benchmarks of multiple strong baselines show that this is a difficult classification task. The results also suggest that simply fine-tuning document-level representations from BERT isn't adequate for this task. Making paragraph-level decisions and aggregating them over the entire document is also ineffective. We present empirical and qualitative analyses that illustrate the specific challenges posed by this dataset. We release~\footnote{Code and dataset released at \href{https://stonybrooknlp.github.io/PerSenT/}{https://stonybrooknlp.github.io/PerSenT/}} 
this dataset with 5.3k documents and 38k paragraphs covering 3.2k unique entities as a challenge in entity sentiment analysis.
\end{abstract}

\section{Introduction}
\blfootnote{
   
    %
    This work is licensed under a Creative Commons 
    Attribution 4.0 International License.
    License details:
    \url{http://creativecommons.org/licenses/by/4.0/}.
}
How can we automatically infer the sentiment of an author towards an entity based only on the text of their news article? This task can be seen as a part of complete document understanding, with potential uses in detecting journalistic bias, and in collecting articles that express certain 
viewpoints towards entities. There are no readily useful datasets or effective solutions for author sentiment inference in the news domain. Sentiment analysis solutions have covered a wide range of domains including movie 
reviews~\cite{pang2008opinion,singh2013sentiment,socher2011semi}, product 
reviews~\cite{dave2003mining,turney2002thumbs,Fang2015}, and social 
media~\cite{abbasi2008sentiment,pak2010twitter}. In the news domain, the closest is inferring the sentiment expressed by one entity (or a group) towards another entity (or a group) mentioned in a news article~\cite{choi2016document}. However, this does not necessarily cover the sentiment expressed by the author towards a specific entity.

To address this gap, we introduce \namep, a crowdsourced dataset of sentiment annotations on news articles about people. For each article, annotators judge what the author's sentiment is towards the main (target) entity of the article. The annotations also include similar judgments on paragraphs within the article. Our experiments with multiple strong classification models show that this is a difficult task that introduces multiple unmet challenges. 

The task we propose is difficult mainly because most information contained in a news article is likely irrelevant for the purposes of inferring the author's sentiment. This is a key difference compared to previously studied sentiment problems in domains such as product or movie reviews, where most information is likely relevant for sentiment inference. On the one hand, a single global document-level representation requires careful aggregation of information related to the target entity that is relevant for sentiment inference. On the other, it is also not easy to make local decisions (say at a paragraph level) and aggregate them --- in many cases paragraph-level decisions can be noisy due to discourse gaps, and furthermore not all paragraph-level decisions should contribute equally to the final decision. Indeed neither approach works satisfactorily. For instance, fine-tuning pre-trained BERT~\cite{devlin2018bert} base model
only yields an F1 of 0.48 on one test set and 0.42 on the more difficult test set for the document-level task. Paragraph-level prediction using BERT is even more difficult, with 0.43 and 0.40 F1 values for the two test sets respectively (see section 3.4 for more details). Motivated by the nature of the task --- requiring focus towards the target entity and aggregating information across the discourse --- we also benchmark models that focus on entity-oriented representations (Recurrent Entity Network), discourse-based representations. However, we find that their performance is worse than the pre-trained BERT classifier.

\begin{figure}[t]
    \centering
    \includegraphics[scale=0.45]{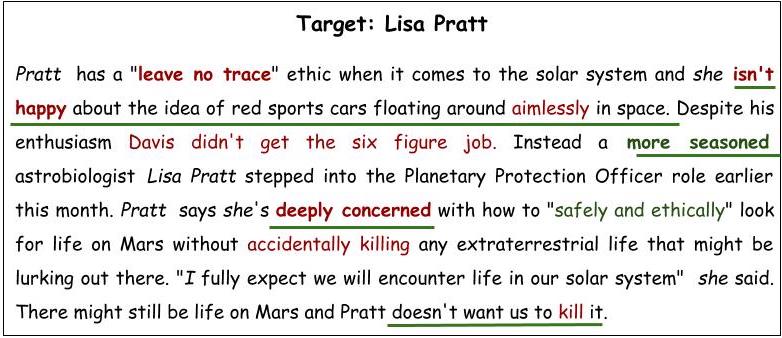}
    \caption{A positive (truncated) document towards the target entity 
    \textit{Lisa Pratt}. A bold green/red word is lexically positive/negative 
    towards the entity. A green/red unbold word is positive/negative towards 
    other entities in the document. The solid green line means the author's view 
    towards the entity is semantically positive on the span. The target entity 
    and all her mentions are in \textit{italics}.}
    \label{fig:task}
    \vspace{-1em}
\end{figure}

We conduct further analysis to show the nature of the challenges introduced by this news domain dataset. One of the main challenges in this task is as shown in Figure~\ref{fig:task}, not all paragraphs necessarily convey the author's sentiment towards the target entity. Thus, making a single decision from a representation of the entire document can be ineffective. Furthermore, our analysis shows that prediction in documents with many unique entities is more difficult. We also find that it is easier to predict document-level sentiment when there are many paragraphs that convey the same sentiment. Last, we also provide a qualitative error analysis that indicates the main categories of inference errors. In summary, this work introduces a new sentiment inference task in the news domain that introduces new challenges in document-level sentiment inference along with a dataset that will support further research in this area.

\section{\namep Dataset}
\label{dataset}

Given a news article about a person (we refer to this as the main or target entity), the author sentiment prediction task is to infer the authors' view towards the main entity. This task requires models to carefully track information that is specific to the entity and aggregate sentiment related information across multiple mentions of the entity in the document. 

Since there are no datasets that directly support this task of author sentiment prediction, we create a new dataset \namep. It consists of news articles that have been previously used for other types of sentiment analysis and related tasks. The criteria for choosing a document is that the news article should mainly talk about a person entity and that it should not be very long or short. We used three resources to select news articles:
\begin{itemize}
    \item MPQA~\cite{deng2015mpqa,wiebe2005annotating}: This dataset contains news articles manually annotated for opinions, beliefs, emotions, sentiments, speculations, etc. It also has target annotations which are entities and event anchored to the heads of noun or verb phrases. All decisions on this dataset are made on sentence-level and over short spans. 
    \item KBP Challenge~\cite{ellis2014overview}: This resource contains TAC 2014 KBP English sentiment slot filling challenge dataset. This is a document-level sentiment filling dataset. In this task, given an entity and a sentiment (positive/negative) from the document, the goal is to find entities toward which the original entity holds the given sentimental view. We selected documents from this resource which have been used in the following similar work in sentiment analysis task~\cite{choi2016document}. 
    \item Media Rank~\cite{ye2019mediarank}: This dataset ranks about 50k news sources along different aspects. It is also used for classifying political ideology of news articles~\cite{kulkarni2018multi}. 
\end{itemize}

We used the following pre-processing steps on the collected news articles. First we find all the person entities in each article, using Stanford NER (Name Entity Resolution) tagger~\cite{finkel2005incorporating} and all mentions of them using co-reference resolution~\cite{clark2016improving,neuralcoref}. 
We removed articles which are not likely to have a main entity of focus. We used a simple heuristic of removing articles in which the most frequent person entity is mentioned only three times or less (even when counting co-referent mentions). For the articles that remain we deemed the most frequent entity to be the main entity of the article. We also filtered out extremely long and extremely short articles  to keep the articles which have at least 3 paragraphs and at most 16 paragraphs. 

Documents are randomly separated into train, dev, and two test sets. We ensure that each entity appears in only one of the sets. Our goal here is to avoid easy to learn biases over entities. To avoid the most frequent entities from dominating the training or the test sets, we remove articles that covered the most frequent entities and use them as a separate test set (referred to as frequent test set) in addition to the randomly drawn standard test set.

\subsection{Crowdsourcing Sentiment Annotations}

We obtained document and paragraph level annotations with the help of Amazon Mechanical Turk workers.
The workers first verified if the target entity we provide is indeed the main entity in the document. 
Then, they rated each paragraph in a document that contained a direct mention or a reference to the target entity. Last, they rated the sentiment towards the entity based on the entire document. In both cases, the workers made assessments about the authors view based on what they said about the target entity. For both paragraph and document level sentiment, the workers chose from five rating categories: Negative, Slightly Negative, Neutral, Slightly Positive, or Positive. We then combine the fine-grained annotations to obtain three coarse-grained classes Negative, Neutral, or Positive. 

\subsection{ Quality Control}
 To ensure quality control, we designed  various assessments. First, at random, we introduced one document for which the sentiment was already rated by an expert. We removed annotations from users whose ratings differed from the expert's rating on this control subset. Second, the workers were required to annotate all paragraphs, to ensure that they read all paragraphs before making decisions about the document. Third, we evaluated different annotations for each document. 
 If all three were different from each other, we reassigned it to two other annotators. Finally, 10\% of annotations were randomly evaluated by experts and  the low-quality annotators are disqualified and their HITs are reassigned to other annotators.
 
 We measured the inter-rater reliability of annotators using weighted Fleiss's Kappa ~\cite{marasini2016assessing}, in which the penalty for each dissimilar classification is based on the distance between two classes. For example, if two annotators classify a document as a positive, the agreement weight is 1, but if one classifies as a positive, and the other classifies as slightly positive the agreement weight is 0.9.  The weights between different classes are shown in Table~\ref{tab:weightsFleis} where negative, slightly negative, neutral, slightly positive, and positive classes are shown with -2, -1, 0, 1, and 2. The weighted agreement number for the whole dataset is 0.79 for paragraph-level annotations and 0.81 for document-level annotations which shows the annotations are highly reliable. Also in 90\% of the instances, at least two out of three annotators agree with each other. For evaluating whether the main entity is extracted correctly or not in each article, if at least two out of three annotators declare that the main entity is not selected correctly, then the document is not included in the dataset. 
 
 \begin{table*}[]
     \centering
     \begin{tabular}{c||c|c|c|c|c|}
          &-2&-1 & 0 & 1 & 2 \\ \hline
         -2 &1&$cos(\pi/8)$&$cos(\pi/4)$&$cos(3\pi/8)$&0\\
         -1 &$cos(\pi/8)$&1&$cos(\pi/8)$&$cos(\pi/4)$&$cos(3\pi/8)$\\
         0  &$cos(\pi/4)$&$cos(\pi/8)$&1&$cos(\pi/8)$&$cos(\pi/4)$\\
          1 &$cos(3\pi/8)$&$cos(\pi/4)$&$cos(\pi/8)$&1&$cos(\pi/8)$\\
         2  &0&$cos(3\pi/8)$&$cos(\pi/4)$&$cos(\pi/8)$&1\\
     \end{tabular}
     \vspace{-0.5em}
     \caption{Inter class weights used for computing inter annotated agreement.}
     \label{tab:weightsFleis}
 \end{table*}{}

\subsection{ Dataset Characteristics}
Few popular entities tend to dominate news collection. In our collection, there were four entities which were the main entity in nearly 800 articles. To avoid these entities from dominating the train or test splits, we moved them to a separate test collection. We split the remaining into a training, development (dev), and test sets at random. Thus our collection includes one standard test set consisting of articles drawn at random (Test Standard), while the other is a test set which contains multiple articles about a small number of popular entities (Test Frequent). The dataset statistics are shown in Table~\ref{dataset-separation}. The distribution among three classes, negative, neutral, and positive is unbalanced: 51\% of documents are positive, 37\% are neutral, and only 11\% are negative. The distribution in paragraph annotations is also unbalanced: 43\% of paragraphs are positive, 46\% are neutral, and only 11\% are negative. Since the documents are a true random sample from news articles, we do not modify these distributions.
\begin{figure*}[t!]
    \centering
    \begin{subfigure}[t]{0.45\textwidth}
        \centering
        \includegraphics[width=.85\textwidth]{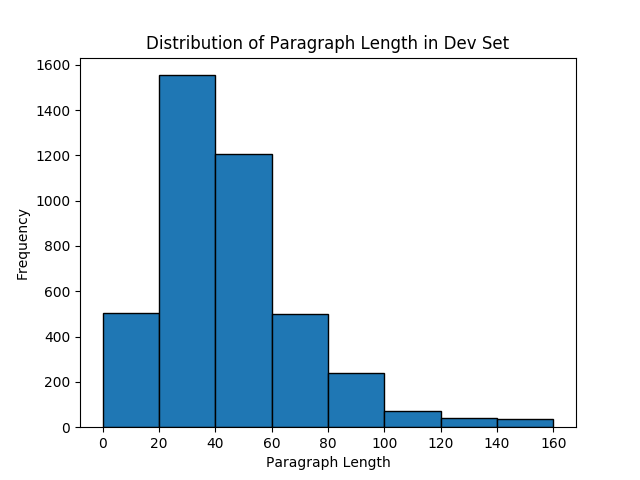}
        \vspace{-0.5em}
         \caption{Distribution of paragraph length in dev set}
        
         \label{fig:par_length_dev}
    \end{subfigure}%
    ~ 
    \begin{subfigure}[t]{0.45\textwidth}
        \centering
       \includegraphics[width=.85\textwidth]{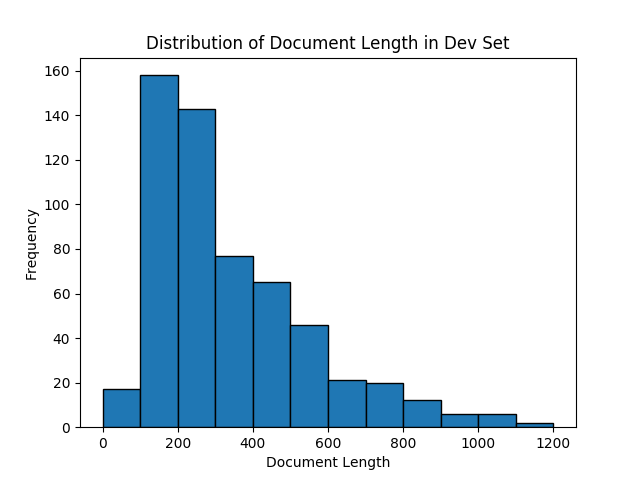}
       \vspace{-0.5em}
    \caption{Distribution of document length in dev set}
    
    \label{fig:doc_length_dev}
    \end{subfigure}
    \vspace{-0.5em}
    \caption{Distribution of paragraph and document length in dev set }
\end{figure*}

\begin{figure*}[t!]
    \centering
    \includegraphics[width=.45\textwidth]{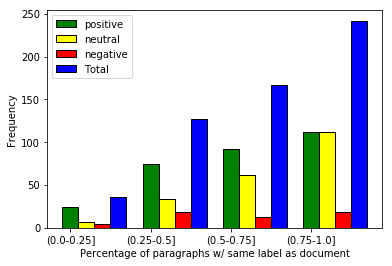}
    \vspace{-1em}
    \caption{Distribution of paragraph and document label agreement}
    
    \label{fig:par_doc_agg_freq}
\end{figure*}{}


Figure~\ref{fig:par_length_dev} shows the distribution of paragraph length (number of words in the paragraph) in the development (dev) set. Most paragraphs are between 20 to 60 words long and very few paragraphs are longer than 128 words long, which use as the maximum length in our experiments. 
Figure~\ref{fig:doc_length_dev} shows the distribution of document lengths (number of words in the document) in terms of the number of words in the development set. As shown in this figure, most of the documents are between 100 to 600 words with only a few larger ones. Based on this distribution, we set the maximum length of documents to 512 in our experiments and trim longer documents~\footnote{In more than 90\% of documents the 512 limit gets us at least 75\% of the document content.}. Overall, the task requires handling long input contexts.

Figure~\ref{fig:par_doc_agg_freq} shows the percentage of paragraphs with paragraph-level and document-level label agreement. As we see more than 70\% of the documents (409 out of 578 documents in dev set) have at least 50\% of label agreement between paragraph labels and document label.

\begin{table*}[t]
\begin{center}
\begin{tabular}{|l|cccc|cccc|c|}
\hline \multirow{2}{*}{Dataset}  &  \multicolumn{4}{c|}{Documents} &   \multicolumn{4}{c|}{Paragraphs}  &  Unique   \\
 {} & Negative & Neutral & Positive & Total   & Negative & Neutral & Positive & Total  & {Entities} \\ 
\hline  Train &351   &1246 & 1758 &3355&  2398&10933&10622& 23953 &  2376  \\
  Dev & 58   &216 & 304&578 &388 &1997&1784& 4169 &  417\\
Test St. &  73&213 &293 &579 &  536&1934&1791& 4261  &  426 \\
  Test Fr. &  139&320&368&827 & 964 &2857&2148& 5969  & 4  \\ 
 \hline
 Total & 621 & 1995 & 2723 & 5339 &4286&17721&16345& 38352 & 3223 \\ 
 \hline
\end{tabular}
\end{center}
\vspace{-1em}
\caption{\label{dataset-separation}We split the dataset into train, dev, and two test sets. Test Standard (Test St.) is a random collection of news articles, whereas Test Frequent (Test Fr.) is a collection that contains multiple articles on a smaller subset of frequently discussed entities.}
\end{table*}

\section{Evaluation}
We benchmark the performance of strong classification baselines to understand and highlight the difficulty of this task.


\subsection{Baselines}
\noindent{\bf Deep Averaging Network (DAN)} 
We use the deep unordered text classification method~\cite{iyyer2015deep} as one of the baselines. 
The method can be seen as a deep neural bag of the words model in which each layer magnifies meaningful differences in the word embedding average, which provides a simple yet strong document-level classification baseline.

\noindent{\bf Modified Recurrent Entity Network}
The original Recurrent Entity Model ~\cite{henaff2016tracking} is equipped with a dynamic, fixed-size memory which can maintain and update a representation of the world. We re-purpose this model for inferring entity sentiment. Firstly, we use two memory slots, one to maintain the information related to the target entity and another to maintain all other information of the world. We initialize the key of the target entity with the corresponding word embeddings denoting the entity. Secondly, we use a bi-directional LSTM to extract sentence-level information, rather than directly updating the memory with word-vectors. This baseline is aimed at testing the benefits of using entity-focused representations.\\
\noindent{\bf Hierarchical Discourse-based LSTM}
Hierarchical sequence models can create multi-level representations over discourse structures~\cite{son2018causal}. We use discourse structures for our sentiment analysis model inspired by prior work that found them useful for related text categorization tasks \cite{ji2017neural}. We extract elementary discourse units (EDUs) using \cite{ji2014representation}, then feed them to Hierarchical BiLSTM models as described in \cite{son2018causal}. This model will encode discourse information for building document-level representations. \\
\noindent{\bf Large Pre-trained Transformer: BERT + Fine-tuning} We use the Bidirectional Encoder Representation from Transformers (BERT)~\cite{devlin2018bert} which produced state of the art results for many NLP tasks. We use the BERT based model (lower-case) and one non-linear layer on top of that for classification. For the document-level task, we set the maximum sequence of the model to 512\footnote{The experiments include documents whose length are greater than 512 see subsection 3.5 for more detailed information} and for paragraph-level experiments we set the sequence length to 128.

\subsection{Prediction over Document-Level Representations}







The first block in Table~\ref{tab:results} shows the performance of the different models for the document-level task, when making predictions over a single representation computed over the entire document. Fine-tuning BERT on this target task performs the best followed by the simpler deep averaging network (DAN) . The more complex Modified REN and Discourse HLSTM models barely outperform a weighted random baseline, showing that complex models cannot learn from scratch effectively on this dataset. More importantly, even the pre-trained BERT model only achieves modest performance overall illustrating the difficulty of this task and the challenge in learning effective document-level representations. The results also show the difficulty of classification for the documents with frequent entities. 
\eat{
\begin{table}[h!]
    \centering
    \begin{tabular}{|c|c|c|}
\hline Model & Test Stand.  &   Test Freq. \\ \hline

Weighted Random &0.33 &0.33 \\ 
Modified REN && \\ 

Discourse HLSTM& 0.38&0.34 
\\ 
BERT Basic & {\bf 0.43}& {\bf 0.41}\\ \hline

    \end{tabular}
    \caption{Macro-F1 on two different test sets using document-level supervision. }
    \label{tab:doc-level}
\end{table}
}
\subsection{Aggregating Paragraph-Level Decision}
Next we describe the performance when making paragraph-level decisions and aggregate them to yield a document-level decision. Table~\ref{tab:results}, second block, shows that when BERT classifier is trained to make paragraph-level decisions that are then directly aggregated (using majority or average), it performs worse than making a single decision over the entire document. We hypothesize that this is mainly because the paragraph level decisions can be challenging because of the limited context (BERT achieves macro F1 of 0.40 in paragraph-level sentiment analysis). 

\subsection{Fine-Tuning for Focused Representations}
\label{sec3-4}
We hypothesize that learning which pieces of information relate to the target entity and which pieces are relevant for sentiment inference can help produce more effective representations. In the extreme case, if the models know exactly which paragraphs to focus on, then they can be more effective at the document-level task. We conducted an oracle experiment, where the paragraphs were annotated with their true paragraph-level sentiment label, and then fed to the document-level BERT. The first row in the third block of Table~\ref{tab:results} (BERT Gold Label) shows the results of this oracle for the test set. 
This suggests that learning from paragraph-level decisions can help document-level prediction if we can learn to make good paragraph-level decisions. 

BERT representations are learned on general language modeling tasks, which may not capture this inductive bias. We devised one fine-tuning task and one multi-task to focus BERT representations:

(i) {\em BERT + ELM Sequential:}  Here the BERT model is first fine-tuned on the paragraph-level prediction task using paragraph supervision, and then further fine-tuned for the document-level task. The results are in BERT Sequential row. This fine tuning results in the best macro-F1 between the paragraph-level supervision models, showing the benefits of focusing the representations on different levels.

(ii) {\em BERT + ELM MultiTask:} We devise an entity-based language modeling task (masked entity language model or Masked ELM), following BERT's original Masked LM task~\cite{devlin2018bert}. For each document in the \namep dataset all entities and their mentions are masked one at a time. Then we use pretrained BERT and retrain it with a classification task to predict whether the masked entity is the main entity of the document or not. For instance, if we have 10 entities and their mentions in a document, we create 10 different versions of that document in Masked ELM dataset. If the masked entity is the main entity, that document is labeled True, otherwise it is labeled as False. Statistics of this Masked ELM task are shown in Table~\ref{tab:mask_entity_stats}. The results of fine-tuning BERT on this Masked ELM task on development set is shown in Table~\ref{tab:masklm}. This task forces BERT to learn contextual information that relate to the target entity token. Since this task is closer to the BERT's original LM task, fine-tuning top 1,2,3, and 4 layers are explored and the best is reported. 
We use this task in two ways. In the sequential mode, we use this task as an intermediate training task, before fine-tuning on the target document-level task. In the multitask mode, we use this ELM task to derive an additional loss. The model is optimized on a combination of both target loss and this additional loss. Neither mode is better than the direct fine-tuning in terms of macro. 
While the initial results here are not positive, integrating multiple losses when fine-tuning models can be tricky and requires more systematic experimentation.




\begin{table}[]
    \centering
    \begin{tabular}{c|c|c|c}
        Set & True Labels & False Labels & Total \\ \hline \hline
         Train&29916&16028& 45944  \\
         Dev &4626&2561&7187 \\
    \end{tabular}
    \vspace{-0.5em}
    \caption{Masked Entity Language Model Dataset Statistics}
    
    \label{tab:mask_entity_stats}
\end{table}



\begin{table}[]
    \centering
    \begin{tabular}{c|c|c|c|c}
        &precision & recall& f1-score &support\\ \hline
         
       False& 0.90&0.72& 0.80 &2561 \\
         True& 0.86& 0.96& 0.91&4626\\\hline 
          macro avg& 0.88&0.84 &0.85& 7187 
         
    \end{tabular}
    \vspace{-1em}
    \caption{Masked ELM on dev set results}
    
    \label{tab:masklm}
\end{table}{}




\begin{table*}[t!]
\begin{center}
\begin{tabular}{|cc|cc|cccc|cccc|}
\hline \multicolumn{2}{|c|}{}& \multicolumn{2}{|c|}{\multirow{2}{*}{Model}} & \multicolumn{4}{c|}{Test Standard}   &  \multicolumn{4}{c|}{ Test Frequent}  \\ 
&&&&Neg&Neut&Pos&MacroF1 &Neg&Neut&Pos&MacroF1  \\
\hline
\multirow{6}{*}{\rotatebox[origin=c|]{90}{Document}}& \multirow{6}{*}{\rotatebox[origin=c|]{90}{Supervision}}&
\multicolumn{2}{|c|}{Weighted Random} &  0.10& 0.30& 0.60& 0.33& 0.11&0.32&0.56&0.33 \\

{}&{}&\multicolumn{2}{|c|}{REN} &0.06&0.34&0.62&0.34&0.04&0.36&0.56&0.32 \\ 

{}&{}&\multicolumn{2}{|c|}{Modified REN} &0.13&0.38&0.56&0.36&0.10&0.45&0.46&0.34 \\

&&\multicolumn{2}{|c|}{Discourse HLSTM}&0.22&	0.43&	0.59&	0.41 &0.18&	0.45&	0.5&0.38 
\\ 

&&\multicolumn{2}{|c|}{DAN} &0.18 &0.50&0.65&0.44&0.11&0.46&0.51&0.36
\\

{}&{}&\multicolumn{2}{|c|}{BERT Classifier} &0.28&0.54&0.62&0.48&0.25&0.54&0.46 &0.42\\ \hline \hline

\multirow{2}{*}{\rotatebox[origin=c|]{90}{Para}}& \multirow{2}{*}{\rotatebox[origin=c|]{90}{Aggr}} &\multirow{2}{*}{BERT}& Majority &0.18&0.52&0.56&0.42&0.29&0.51&0.44&0.41 \\

&{}&&Average &0.16&0.51&0.56&0.41&0.28&0.52&0.46&0.42 \\  \hline \hline

\multirow{3}{*}{\rotatebox[origin=c|]{90}{Foc}}& \multirow{3}{*}{\rotatebox[origin=c|]{90}{Rep}}&\multicolumn{2}{|c|}{BERT Gold Label} &0.82&0.78&0.87&0.82&0.79&0.74&0.82&0.78\\ 

& &\multicolumn{2}{|c|}{BERT+ELM Sequential} &0.12&0.52&0.66&0.43&0.16&0.48&0.51&0.38 \\

&&\multicolumn{2}{|c|}{BERT+ELM MultiTask} &0.17&0.48&0.60&0.42&0.24&0.47&0.49&0.40 \\ 
\hline

\end{tabular}
\end{center}
\vspace{-1em}
\caption{ Macro F1-score for different models on each class. There are three main classes: Negative (Neg), Neutral (Neu), and Positive (Pos). For each class F1 is reported and for each model macro F1 is reported as it is very important for the model to predict the minority class (Negative) as well as the other two classes. The model which has higher Macro-F1 on Dev set is selected and the performance on two test sets are evaluated. ELM is the short form of ELM which is discussed in \ref{sec3-4}}

\label{tab:results}
\end{table*}

\subsection{Analysis}

\begin{figure}[t!p]
    \centering
    \begin{subfigure}[b]{0.48\textwidth}
        \centering
      \includegraphics[width=.88\textwidth]{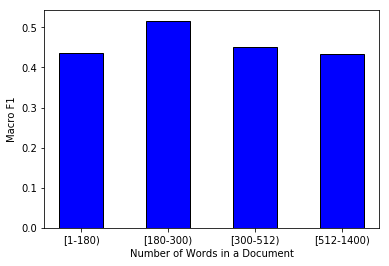}
      \vspace{-.8em}
    \caption{\small F1-score vs the number of words in a document for document-level BERT  }
    
    \label{fig:avgF1}
    \end{subfigure}%
\hfill
\begin{subfigure}[b]{0.48\textwidth}
    \centering
    \includegraphics[width=.88\textwidth]{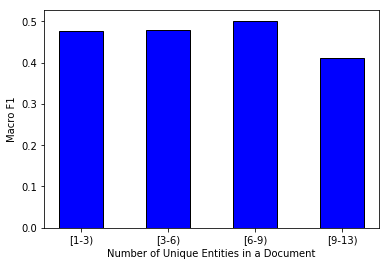}
    \vspace{-.8em}
    \caption{\small F1-score for different number of unique entities in a document }
    \label{fig:uniqEnt}
\end{subfigure}%

\hfill
    \begin{subfigure}[b]{0.48\textwidth}
        \centering
        \includegraphics[width=.88\textwidth]{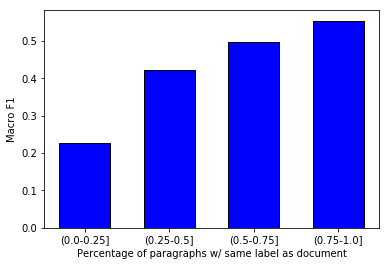}
        \vspace{-.8em}
    \caption{\small F1-score for different percentage of agreements between document level label and paragraph labels }
    \label{fig:aggr}
    \end{subfigure}{}
\hfill
\begin{subfigure}[b]{0.48\textwidth}
    \centering
    \includegraphics[width=.88\textwidth]{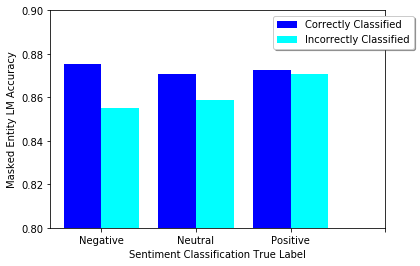}
    \vspace{-.8em}
    \caption{\small Comparison of Masked ELM accuracy with sentiment classification task }
    
    \label{fig:masklmsent}
\end{subfigure}{}

\vspace{-0.5em}
    \caption{Analyzing F1 score in different criteria}

\end{figure}

\noindent{\bf Document Length} Figure~\ref{fig:avgF1} shows how model performance varies with document length. Model performance is highest for documents length between 180-300 words, the length that is most prevalent in the training data. For shorter and longer documents the performance is significantly lower, hinting at the difficulty in generalizing along the length dimension. For documents longer than 512, which is the input size limit for BERT, the performance does not drop too much. This is likely because we have trimmed longer documents in the training data to 512 tokens. Using models without this length limitation (e.g., XLNet~\cite{DBLP:journals/corr/abs-1906-08237} and Longformer~\cite{Beltagy2020Longformer}) should help address this issue directly. \\
\noindent{\bf Para vs. Document label agreement} Figure~\ref{fig:aggr} shows the importance of agreement between document level labels and paragraph level labels. The higher the agreement between document and paragraph labels, the higher the macro F1 score. It can be a good test of generalization to see if models can assess document-level sentiment accurately even in the presence of conflicting local paragraph-level sentiments.

\noindent{\bf Number of entities} We hypothesized that if there are more entities in a document then the models will fail because they will have a tough time aggregating information relevant to the target entity. Figure~\ref{fig:uniqEnt} shows the macro F1 score vs the number of unique entities in a document. It turns out that only when the number of unique entities is substantially high (more than 10) there is a significant drop in performance. For smaller number of entities the trend is in the opposite direction.\\
\noindent{\bf Target Entity Failure} We expect the classification to fail when the model is unable to reliably predict information about the target entity. The Masked ELM task is a proxy for this test. Figure~\ref{fig:masklmsent} shows the relation between the performance of BERT on the sentiment classification and the Masked ELM tasks. When BERT is able to reliably locate the target entity, i.e., when it has higher accuracy in Masked ELM task, it is usually also more accurate in classifying the sentiment towards that entity. This further motivates better exploration tasks that ensure focusing representations towards the target entity.

\begin{figure}
    \centering
    \includegraphics[width=.5\textwidth]{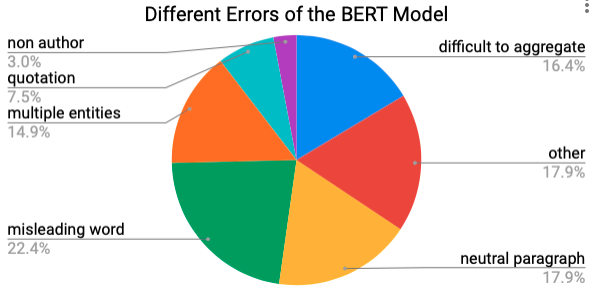}
    \caption{\small Different types of errors for BERT classifier on dev set. }
    \vspace{-1em}
    \label{fig:mistake_type}
\end{figure}

\noindent{\bf Error Analysis} We conducted a manual error analysis to identify the type of mistakes. We analyzed a random subset of 10\% of the documents from development set which are misclassified by BERT. The results are shown in Figure~\ref{fig:mistake_type}. We see misleading words, neutral paragraphs, and long documents are most prevalent in documents where BERT fails. Other minor categories include cases where much of non-author sentiment is expressed, presence of quotations and multiple entities. The other category combines errors in finding the main entity, error in annotation, and other unknown reasons.

    

\section{Related Work}

\subsection{ Aspect-Level Sentiment Analysis}
The goal is to determine the sentiment toward specific aspect of a target such as product, movie, restaurant, etc~\cite{pang2008opinion,schouten2016survey}.  While there are broad similarities, the main objective here is to infer the author's view towards an entity and not towards different "aspects" of the entity. 
In \cite{chen-etal-2017-recurrent-attention}, the authors proposed a multiple attention mechanism to capture sentiment features and then combined the result with a recurrent neural network. By using the weighted memory network, they captured the sentiment features automatically and used its capacity to capture sentiments toward different opinion targets of a sentence. They evaluate their model on four different sentence level target specific sentiment analysis and outperforms previous models. \newcite{ma2018targeted} proposed a novel model by augmenting the long short term memory (LSTM) network with a hierarchical attention mechanism consisting of a target level attention and a sentence-level attention. Commonsense knowledge is also incorporated into the training for sentiment classification as an external knowledge. The major difference of aspect level sentiment analysis and our task is that, in aspect-level the text is talking about different aspects of one element and the goal is to find the sentiment towards each aspect. The aspect-level reviews are for short texts and the opinion towards each aspect is persistent in the text. In entity-based document-level sentiment analysis the goal is to find the whole author's view towards one entity, not different aspect of one entity. 
{}
\subsection{Document-Level Sentiment Analysis} 
Many works have studied document-level sentiment analysis~\cite{pang2002thumbs,moraes2013document,lin2006side,tang2015document}.
~\newcite{mcdonald2007structured} showed the effectiveness of sentence level prediction on document level classification. This paper focuses on models to jointly classify sentiment on multiple levels of granularity, i.e. on sentence level and document level. The authors try to show how using the information on single sentences can help the sentiment prediction task on the whole document. 
Our work is inspired by recent efforts that focus on entity-centric sentiment model, which learnt sentiments between pairs of entities~\cite{choi2016document}. While closely related this task cannot be re-purposed to infer author sentiment directly. 
\subsection{News Domain Sentiment Analysis }Sentiment analysis in news articles has been studied from various perspectives for different types of applications~\cite{yu2013impact,ahmad2007multi,godbole2007large,balahur2013sentiment}. We propose a new dataset and task that requires different levels of linguistic features (i.e., discourse units, sentences, and entities) for accurate sentiment inferences.

\subsection{Stance Detection}
Stance detection is the task of detecting whether the author is in favor of the target or against it and it is important to determine not only the author's stance expressed in her/his debate posts but also the reasons why s/he supports or opposes the issue under debate \cite{hasan2014you,zarrella2016mitre}. Most of the researches in stance detection field are using debate posts in forums or political congress debates~\cite{walker2012stance,hasan2013stance,augenstein2016stance}. The stance detection texts are usually not longer than a forum post~\cite{hasan2013stance} or twitter text~\cite{mohammad2016semeval}, and the classification task is usually to predict whether the author is in favor or against the target. 
In stance detection the models are looking for favorability towards a given target of interest which may not be mentioned or targeted in the text~\cite{mohammad2016semeval,augenstein2016stance}. But in sentiment analysis the author is explicitly mentioning the target and it is formulated as finding whether the text is positive, negative, or neutral towards the main entity. As such, our task is categorized as sentiment analysis task where the targets are persons and mentioned directly in the text. However, the authors may not explicitly express their opinion.
\section{Conclusions \& Discussion}
In this work, we introduced author sentiment inference in the news domain, a new sentiment task and release a challenge dataset. Most information contained in a news article might be irrelevant for the purposes of inferring the authors sentiment towards the target entity. This makes the task quite challenging for both models that operate over global representation and for models that aggregate local decisions. We hope the task and the dataset will be a useful resource for the community to build more effective text classification models. 

\section{Acknowledgements}
This work was supported in part by the National
Science Foundation under Grant IIS-1815358. We also thank Google for supporting this research through Google Cloud Credits program.

\bibliographystyle{coling}
\bibliography{coling2020}

\end{document}